\documentclass{article}

\usepackage{fullpage}
\date{}

\usepackage{amsmath,amssymb,amsthm,amsfonts}
\usepackage{graphicx}
\usepackage{booktabs}
\theoremstyle{plain}
\newtheorem{theorem}{Theorem}[section]
\theoremstyle{definition}
\newtheorem{definition}[theorem]{Definition}

\usepackage{enumerate}
\usepackage{algorithm}

\providecommand{\keywords}[1]
{
  \small	
  \textbf{\textit{Keywords---}} #1
}

\title{Deblurring Processor for Motion-Blurred Faces Based on Generative Adversarial Networks}

\author{
  Shiqing Fan \\
    School of Informatics\\
  Xiamen University\\
  Xiamen,  Fujian 361005, China \\
   \texttt{loy.fsq@gmail.com} \\
  \and
   Ye Luo\\
 School of Informatics\\
  Xiamen University\\
  Xiamen,  Fujian 361005, China \\
  \texttt{luoye@xmu.edu.cn, luoye80@gmail.com} \\
}

\begin{document}
\maketitle

\begin{abstract}
Low-quality face image restoration is a popular research direction in today's computer vision field. It can be used as a pre-work for tasks such as face detection and face recognition. At present, there is a lot of work to solve the problem of low-quality faces under various environmental conditions. This paper mainly focuses on the restoration of motion-blurred faces. In increasingly abundant mobile scenes, the fast recovery of motion-blurred faces can bring highly effective speed improvements in tasks such as face matching. In order to achieve this goal, a deblurring method for motion-blurred facial image signals based on generative adversarial networks(GANs) is proposed. It uses an end-to-end method to train a sharp image generator, i.e., a processor for motion-blurred facial images. This paper introduce the processing progress of motion-blurred images, the development and changes of GANs and some basic concepts. After that, it give the details of network structure and training optimization design of the image processor. Then we conducted a motion blur image generation experiment on some general facial data set, and used the pairs of blurred and sharp face image data to perform the training and testing experiments of the processor GAN, and gave some visual displays. Finally, MTCNN is used to detect the faces of the image generated by the deblurring processor, and compare it with the result of the blurred image. From the results, the processing effect of the deblurring processor on the motion-blurred picture has a significant improvement both in terms of intuition and evaluation indicators of face detection.
\newline
\\
\normalsize \textbf{CCS CONCEPTS ·} Computing methodologies \textbf{·} Artificial intelligence  \textbf{·} Computer vision
\newline
\end{abstract}

\section{Introduction}\label{1}

\subsection{Motivation}
Face detection\cite{rowley1998neural} and face recognition \cite{zhao2003face} are among the most popular and widely used tasks in the field of computer vision. In practical applications, data sets are often affected by changes in the surrounding environment and scenes, and have different characteristics. Generally speaking, it frequently happens that the raw face images  are not ideal for detection, such as images with some missing parts, of low resolution, or being fogged or blurred. There have already been a lot of researchers proposing much cutting-edge work to help solve problems related to the detection and recognition of low-quality face data. For instance, the work of Yeh \emph{et al.} focused on repairing defective images \cite{yeh2017semantic}, and Chen \emph{et al.} presented an approach about face image super-resolution \cite{chen2018fsrnet}. There is also some research work about low-resolution face recognition \cite{zou2011very}, single image dehazing \cite{yang2018proximal} etc. Under the current increasing demand for mobile face recognition, the detection and recognition of blurred facial images, especially of motion-blurred facial images, have become highly significant. The purpose of this paper is to present a picture-processing algorithm applied to motion-blurred face pictures, providing an enhanced detectability for face detection.

After Generative Adversarial Nets (GANs) was proposed by Goodfellow \emph{et al.} , it has become an exceedingly popular model in deep learning, especially in the field of image generation. These image-generating models based on GAN can afford very natural output, such as face generation \cite{goodfellow2014generative}, face aging \cite{antipov2017face}, image style transfer \cite{isola2017image}, and face beautification \cite{li2018beautygan}. It can be said that GAN generative models have reached the level of being fake but not seen through. 

Applying the GAN model to image processing is also an interesting idea. Traditional generative models get different outputs from noise, while GAN in image processing uses defective pictures in the real world as input, and builds a generator, though which excellent pictures can be obtained. This end-to-end model is very intuitive, a generator that can transform defective pictures to excellent pictures is trained to be an image processor for some pictures that are difficult to be processed by traditional algorithms to. Therefore, the goal of this article is to apply the GAN structure to the motion-blurred face image processing to form a GAN-based motion-blurred face images processor, and conduct a follow-up face detection experimental comparison test on the effectiveness of the processor.

\subsection{Contribution}
In this paper, some changes are made on the basis of the traditional GAN structure and applied to the deblurring processing of the motion-blurred images of human faces. First, we perform motion blur simulation processing on some public data set of face pictures to obtain pairs of motion-blurred and sharp face images, and then we design the generator in the GAN structure to use motion-blurred faces as input and sharp faces as output. The discriminator in the network judges the generated images and the sharp images separately. The generated image and the sharp images have pseudo image tags and real image tags respectively. We used several different GAN loss function designs for specific training. After training, the generator part in our network is the face deblur processor we ultimately need. Finally, we use this processor to test the image processing of face images, and conduct comparative experimental statistics for face detection to test the capabilities of the processor. We found that the face images through the processor has a significant improvement in the success rate of face detection and the positioning accuracy of the position of faces and also the facial features.

We introduced the related work of generative adversarial networks and deblurring technology for image motion blur in Section 2, and introduced some formulaic basic knowledge, including GAN, motion blur and face detection, in Section 3. In Section 4, we give an intuitive explanation and detailed introduction of the network used in this article. Section 5 is about the whole experimental process and effect demonstration. Finally, we give a summary of the paper.

\section{Related Work}\label{2}
\subsection{Image Deblurring}\label{2.1}

The previous early work on the processing of blurred images was mainly non-blind, meaning that the cause of the blur and the specific effect and degree of the blur are known and formulaic expressions can be studied, that is, the blur kernel k in the following Equation 2.1 is known.

\begin{definition}\label{D:blur-kernel}The common blur model can be formulated as
$$I_{B}=k(M) * I_{S}+N$$
where $I_B$ is a blurred image, $I_S$ is a sharp image, $k(M)$ are unknown blur kernels determined by motion field $M$. $N$ is other noise. 
\end{definition}

The formula indicates that the blurred image is formed by the convolution of the blur kernel and the clear image with some noise added. There has been a lot of research work on fuzzy image processing with prior knowledge, that is, the knowledge about the blur kernels. For example, some work relying on the classic Lucy-Richardson algorithm, obtaining $I_S$ estimates by performing deconvolution operations with Wiener or Tikhonov filters. Starting from the success of Fergus \emph{et al.} \cite{fergus2006removing}, many methods have been extended in recent years \cite{xu2013unnatural} \cite{xu2010two}\cite{perrone2014total}\cite{babacan2012bayesian}. These methods assume that the blur is uniform throughout the image and try to compensate for the blur caused by camera shake. They first estimate the motion of the camera based on the blur kernel caused by the scene, and then reverse the effect by performing a deconvolution operation. With the success of deep learning, in the past few years, some methods based on Convolutional Neural Networks (CNNs) have emerged \cite{gong2017motion}\cite{sun2015learning}. All these methods use CNNs to estimate the unknown ambiguity function. Later, some kernelless end-to-end methods appeared. Noorozi \emph{et al.} \cite{noroozi2017motion} and Nah \emph{et al.} \cite{nah2017deep} used multi-scale CNNs to directly blur the images. Ramakrishnan \emph{et al.} \cite{ramakrishnan2017deep} uses a combination of the pix2pix framework \cite{isola2017image} and a densely connected convolutional network \cite{huang2017densely} to perform kernel-free blind image deblurring. The advantage of these new methods is that they can handle different sources of blur. And Kupyn \emph{et al.} \cite{kupyn2018deblurgan} even pioneered the use of Generative Adversarial Networks(GAN) structure for image deblurring to help object detection.

\subsection{Generative Adversarial Networks}\label{2.2}

GAN was proposed by Goodfellow \emph{et al.}\cite{goodfellow2014generative}. It uses the game learning of generative network and adversarial network to finally produce excellent generative output, which has great influence in unsupervised learning of complex distribution. Two models, generator and discriminator, are used in the GAN framework to train at the same time, similar to the relationship between a spear and a shield. During the training process, the discriminator’s discriminatory ability is getting stronger and stronger, and at the same time, the generator’s forgery ability also getting stronger in order to fool the discriminator. In an ideal state, the final result of the game is when the discriminator’s judgment result of the generator’s forged output with a probability of 0.5 is true, which indicates that the generator is sufficient to produce a image that cannot be distinguished from the real image. Then the generative model can be used to other downstream tasks. After the basic GAN model was proposed, many improvements to the details of GAN were also proposed. Mao \emph{et al.} \cite{mao2017least} proposed a least squares generative adversarial networks(LSGANs), using the least squares loss function in the discriminator, the purpose is to solve the problem of gradient disappearance that may exist in traditional GANs to generate more stable and high-quality images. The Wasserstein GAN(WGAN) proposed by Arjovsky \emph{et al.} \cite{arjovsky2017wasserstein} focuses on the balance between generator and discriminator. For example, once discriminator is trained too well, generator cannot learn effective gradients, and WGAN truncates the part of discriminator when updating parameters, that is, "gradient clipping". What’s more, their later work \cite{gulrajani2017improved} improved the model to use gradient penalty.

\section{BASIC CONCEPTS}\label{3}
In this section, we introduce several GAN network ideas in formulaic form, and give the specific form of motion blur discussed in this paper, and finally give some basic concepts of face detection, which provides the basic knowledge beforehand for our GAN-based motion blur facial image processor.

\subsection{GAN Loss}\label{3.1}
In the general GAN structure, we train two models at the same time, one is the generator $G$ and the other is the discriminator $D$. In general tasks, we will have some real data $\boldsymbol{x}$, and some network inputs, such as random noise, set as $\boldsymbol{z}$. The output of $\boldsymbol{z}$ through G in the network is $G(\boldsymbol{z})$. The main task of G is to learn the distribution $p_{\mathrm{data\ }}(\boldsymbol{x})$ of data $\boldsymbol{x}$, its output $G(\boldsymbol{z})$ should be as close to $\boldsymbol{x}$ as possible. On the other hand, the main task of the D is to distinguish between $\boldsymbol{x}$ and $\boldsymbol{z}$, that is, to accurately determine that $\boldsymbol{x}$ is real data and $\boldsymbol{z}$ is forged data. From the above ideas, we can see that the goal of traditional GAN can be formulated as Definition~\ref{D:GAN} . Definition~\ref{D:LSGAN}  shows the form of the LSGANs, which is similar to the traditional one, but with the cross entropy in the traditional GAN changed to the least square loss. As for the WGANs, the Earth-Mover (also called Wasserstein-1) distance $W(q,p)$ is used instead, in order to transform the distribution $q$ into the distribution $p$. With reference to Kantorovich-Rubinstein duality \cite{villani2008optimal}, we get the WGAN value function of Definition~\ref{D:WGAN}(i), and then the improved version WGAN-GP taking into account gradient penalty is shown in Definition~\ref{D:WGAN}(ii).

\begin{definition}\label{D:GAN} 
The original GANs
$$
\min _{G} \max _{D} V(D, G)=\mathbb{E}_{x \sim p_{\text {data }}(x)}[\log D(x)]+\mathbb{E}_{z \sim p_{z}(z)}[\log (1-D(G(z)))]
$$
\end{definition}

\begin{definition}\label{D:LSGAN} 
LSGANs \cite{mao2017least}
$$
\begin{array}{l}
\min _{D} V_{L S G A N}(D)=\frac{1}{2} \mathbb{E}_{\boldsymbol{x} \sim p_{\text {data }}(\boldsymbol{x})}\left[(D(\boldsymbol{x})-b)^{2}\right]+\frac{1}{2} \mathbb{E}_{\boldsymbol{Z} \sim p_{\boldsymbol{z}}(\boldsymbol{z})}\left[(D(G(\boldsymbol{z}))-a)^{2}\right] \\
\min _{G} V_{L S G A N}(G)=\frac{1}{2} \mathbb{E}_{\boldsymbol{Z} \sim p_{\boldsymbol{z}}(\boldsymbol{z})}\left[(D(G(\boldsymbol{z}))-c)^{2}\right]
\end{array}
$$
where a and b are the labels for fake data and real data, respectively. c denotes the value that G wants D to believe for fake data.
\end{definition}

\begin{definition}\label{D:WGAN} 
Wasserstein GANs
\begin{enumerate}[(i)]

\item WGAN \cite{arjovsky2017wasserstein}
$$
\operatorname{minmax}_{G} \underset{D \in \mathcal{D}}{\mathbb{E}}_{\boldsymbol{x} \sim \mathbb{P}_{r}}[D(\boldsymbol{x})]-\underset{\widetilde{\boldsymbol{x}} \sim \mathbb{P}_{g}}{\mathbb{E}}[D(\widetilde{\boldsymbol{x}})]
$$
where $\widetilde{\boldsymbol{x}}=G(z)$, $\mathbb{P}_r$ and $\mathbb{P}_g$ respectively represent the distribution of $\boldsymbol{x}$ and  $\widetilde{\boldsymbol{x}}$, and $\mathcal{D}$ is the set of 1-Lipschitz functions. The purpose of the equation is to minimize $W(\mathbb{P}_r,\mathbb{P}_g)$.

\item WGAN-GP \cite{gulrajani2017improved}
$$
L=\underbrace{\underset{\tilde{\boldsymbol{x}} \sim \mathbb{P}_{g}}{\mathbb{E}}[D(\widetilde{\boldsymbol{x}})]-\underset{\boldsymbol{x} \sim \mathbb{P}_{r}}{\mathbb{E}}[D(\boldsymbol{x})]}_{\text {Original loss }}+\underbrace{\lambda_{\widehat{\boldsymbol{x}} \sim \mathbb{P}_{\hat{\boldsymbol{x}}}} \mathbb{E}\left[\left(\left\|\nabla_{\widehat{\boldsymbol{x}}} D(\widehat{\boldsymbol{x}})\right\|_{2}-1\right)^{2}\right]}_{\text {Gradient penalty }}
$$
where $\hat{\boldsymbol{x}}\sim\mathbb{P}_{\hat{\boldsymbol{x}}}$ is random samples whose gradient will be punished.
\end{enumerate}
\end{definition}

\subsection{Motion-Blurred Image Generation}\label{3.2}
We use the blur kernels generation algorithm proposed by \cite{kupyn2018deblurgan} to get the blurred face image data. This method refers to the random trajectory generation idea of Boracchi \emph{et al.} \cite{boracchi2012modeling}. After generating the trajectory, sub-pixel interpolation is applied to the trajectory vector to generate a blur kernel. This method can simulate a more realistic and complex blur kernel. This is summarized in the Algorithm~\ref{alg:Blur}. In order to test the effect of our deblurring processor, we set a relatively high level of motion blur.

\begin{algorithm}[htb] 
\caption{Motion Blur Image Generation} 
\label{alg:Blur} 
\leftline{Initialize the velocity vector of the trajectory points(1);}

\leftline{Initialize the position of trajectory points to zeros;}

\leftline{\textbf{For} the number of iterations, \textbf{do}}

	Generate the velocity of the next moment(2);
	
	Get the position of the next point according to the current speed;
	
\leftline{\textbf{End}}

\leftline{Get the trajectory vector x;}

\leftline{Generate blur kernel by sub pixel interpolation(x);}

\leftline{Get Blur image by convolution operation of blur kernels and sharp images.}

\leftline{}

\leftline{\emph{(1) According to the previously set maximum movement length and number of iterations.}}

\leftline{\emph{(2)Based on the previous point velocity and position, gaussian perturbation, impulse perturbation and}}

\leftline{\emph{deterministic inertial component.}}
\end{algorithm}

\subsection{Face Detection}\label{3.3}

Face Detection is to find out the position of all faces in an image. It is usually framed by a rectangular frame. The input is an image, and the output is a number of rectangular frame positions like $(x,y,w,h)$ containing faces.

The evaluation indicators for face detection include the following:

\emph{Recall Rate}: Since the number of faces contained in each image is uncertain, it is measured by the ratio of detection. The closer the rectangular frame detected by the detector is to the manually labeled rectangular frame, the better the detection result. Usually, if the intersection ratio is greater than 0.5, it is considered to be detected. So recall = (the number of detected faces)/(the total number of faces in the image).

\emph{False Positives}: We use the absolute number of detection errors to express. In contrast to recall, if the IoU(Intersection over Union) of the rectangular box detected by the detector and any manually labeled box is less than 0.5, the detection result is considered to be a false detection.

\emph{Detection Speed}: Usually expressed by frame-per-second (FPS).

\section{PROCESSOR FOR BLURRED IMAGES OF HUMAN FACES}\label{4}

\subsection{Processor Network Structure}\label{4.1}

\begin{figure}[h]
\vspace{.1in}
\centerline{\includegraphics[scale=0.2]{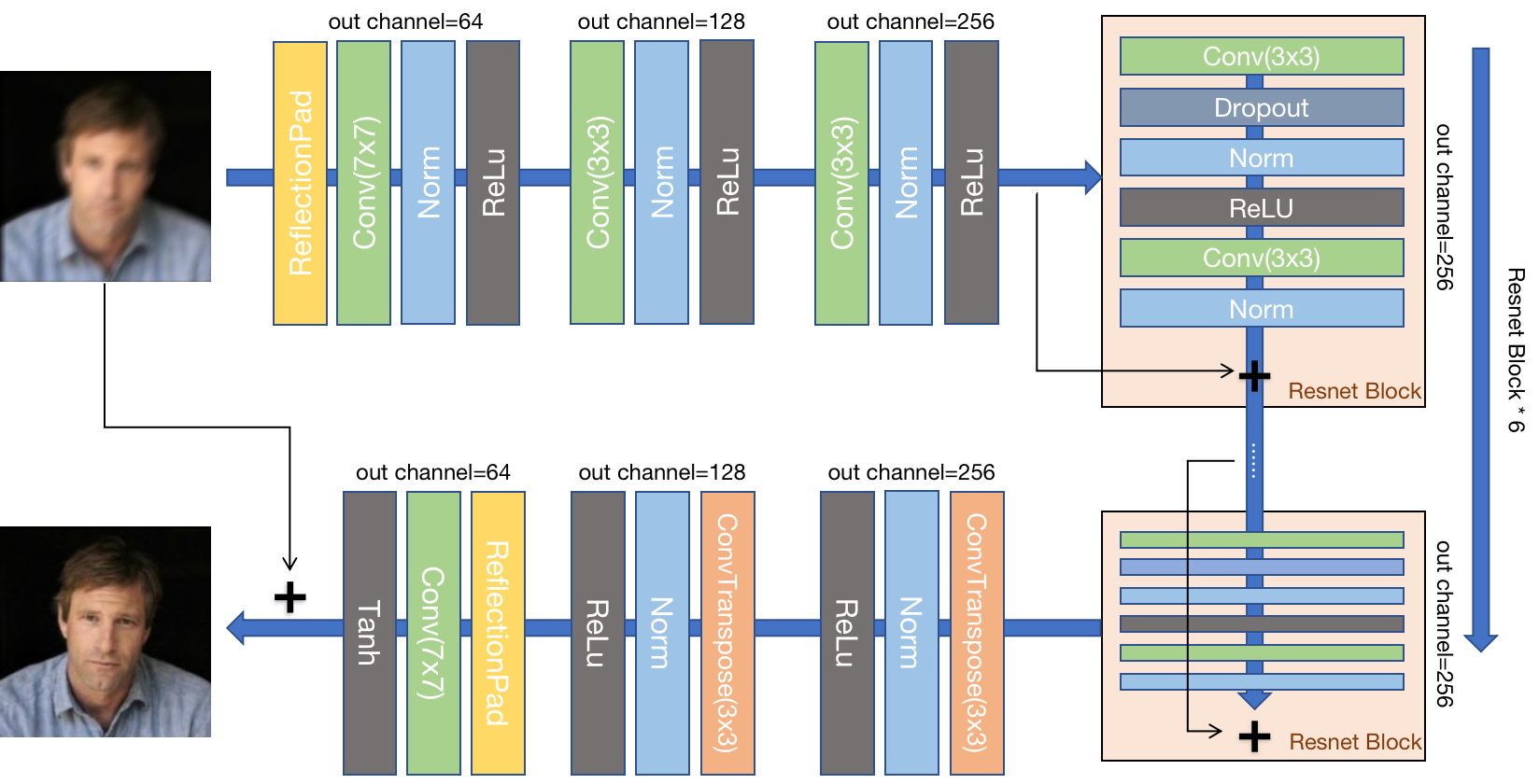}}
\caption{Deblurring Processor for Motion-Blurred Faces(Generator of GAN)}
\label{F:net}
\end{figure}

Our motion face deblurring processor model is shown in Figure~\ref{F:net}. It should be noted that this structure is our final processor model, which is part of the generator in the GAN structure. The generator is composed of an up-sampling part including three convolution blocks, a ResNet \cite{he2016deep} part including 6 ResNet blocks, and a down-sampling part including 2 transposed convolution blocks and the final output convolution block.

\subsection{Model Training}\label{4.2}

As we mentioned above, our GAN structure includes a generator and a discriminator, and they need to be trained at the same time. When training, the input is a pair of blurred and sharp images, model gets the restored images from generator, and both the restored images and the sharp images are input to the discriminator. For the generator, the restored picture is marked as the real picture for the loss calculation. Meanwhile the training of the discriminator uses the calculated value obtained by marking the restored images as fake and the sharp images as real as the loss, so as to achieve the game effect of G and D. We use three GAN types at the same time, that is, three training loss calculation methods mentioned in Subsection~\ref{3.1} for experiments.

\begin{figure}[ht]
\vspace{.1in}
\centerline{\includegraphics[scale=0.2]{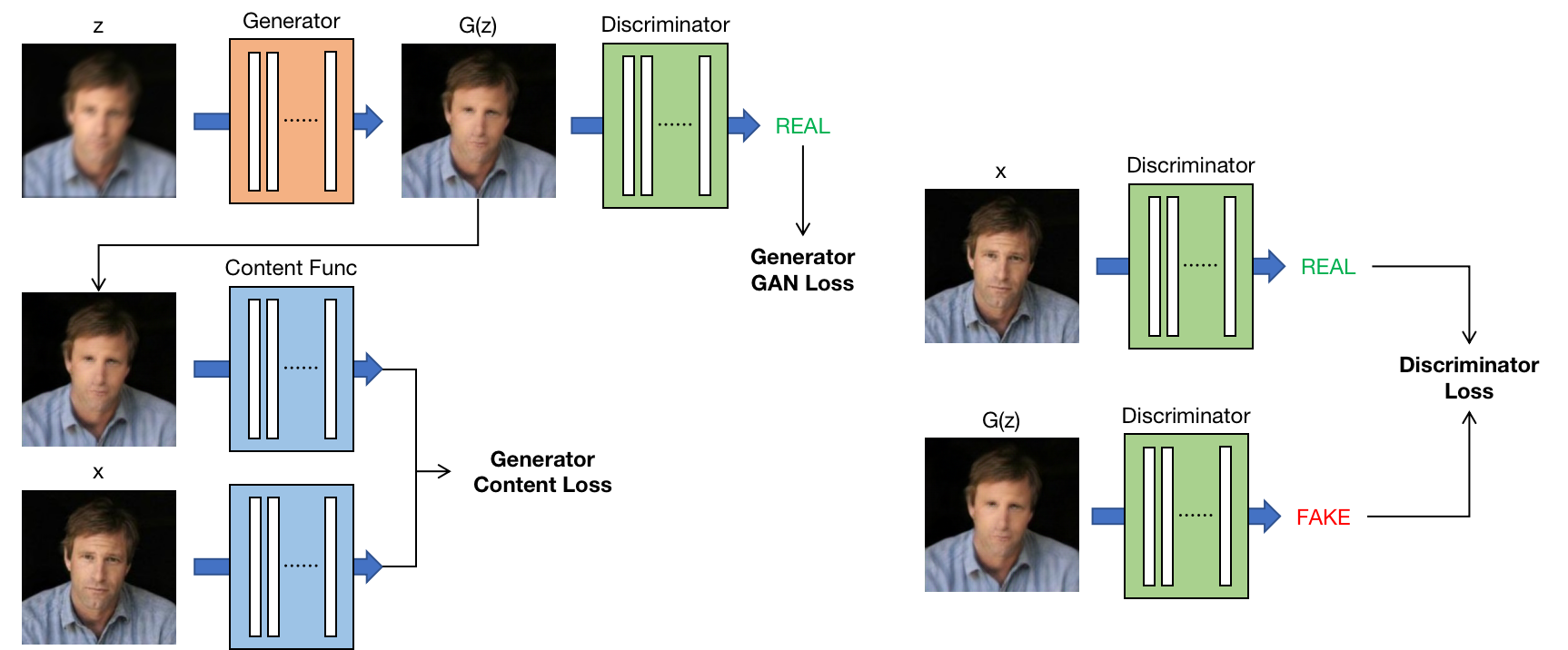}}
\caption{Training a GAN for motion blur face image processing.}
\label{F:loss}
\end{figure}

Specifically, as shown in Figure~\ref{F:loss}, the training loss function of the generator G consists of two parts, one part is the result of the restored images G(z) generated by the generator after passing the discriminator, and the loss calculated by the label REAL, which is used to improve the forgery ability of the generator. The other part is the content loss between restored images $\boldsymbol{z}$ and sharp images $\boldsymbol{x}$, they first pass through a content function(similar to VGG network\cite{simonyan2014very}) and then calculate the difference, which is also to improve the generator's ability to generate images that are more similar to sharp images. While the loss function of the discriminator is calculated from the discriminant results of $\boldsymbol{x}$ and $G(\boldsymbol{z})$, where the discriminant label of the picture is real.

\section{EXPERIMENT DISPLAY}

\subsection{Motion-Blurred Face Data Set}\label{5.1}

\begin{figure}[ht]
\vspace{.1in}
\centerline{\includegraphics[scale=0.2]{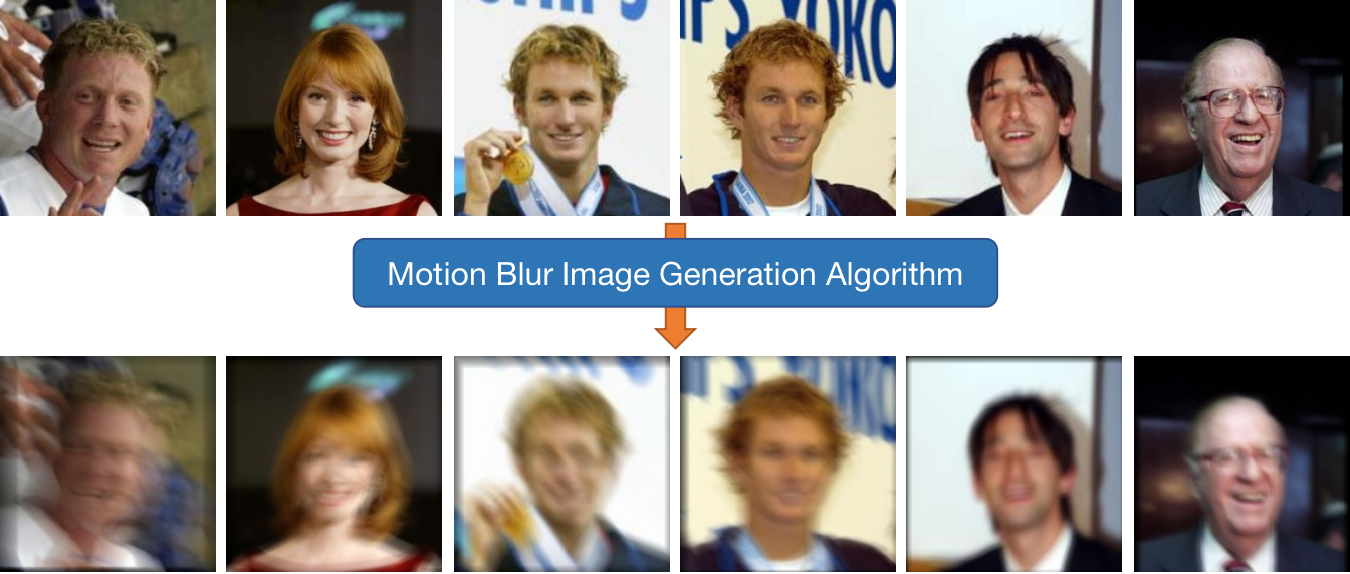}}
\caption{Display of some sharp images and motion-blurred images of human faces.}
\label{F:blur}
\end{figure}

We used the LFW (Labeled Faces in the Wild Home) \cite{huang2008labeled} dataset for experiments. Figure~\ref{F:blur} shows some examples of motion-blurred face images obtained by Algorithm~\ref{alg:Blur}.

\subsection{Image Processing and Face Detection}

\begin{figure}[ht]
\vspace{.1in}
\centerline{\includegraphics[scale=0.2]{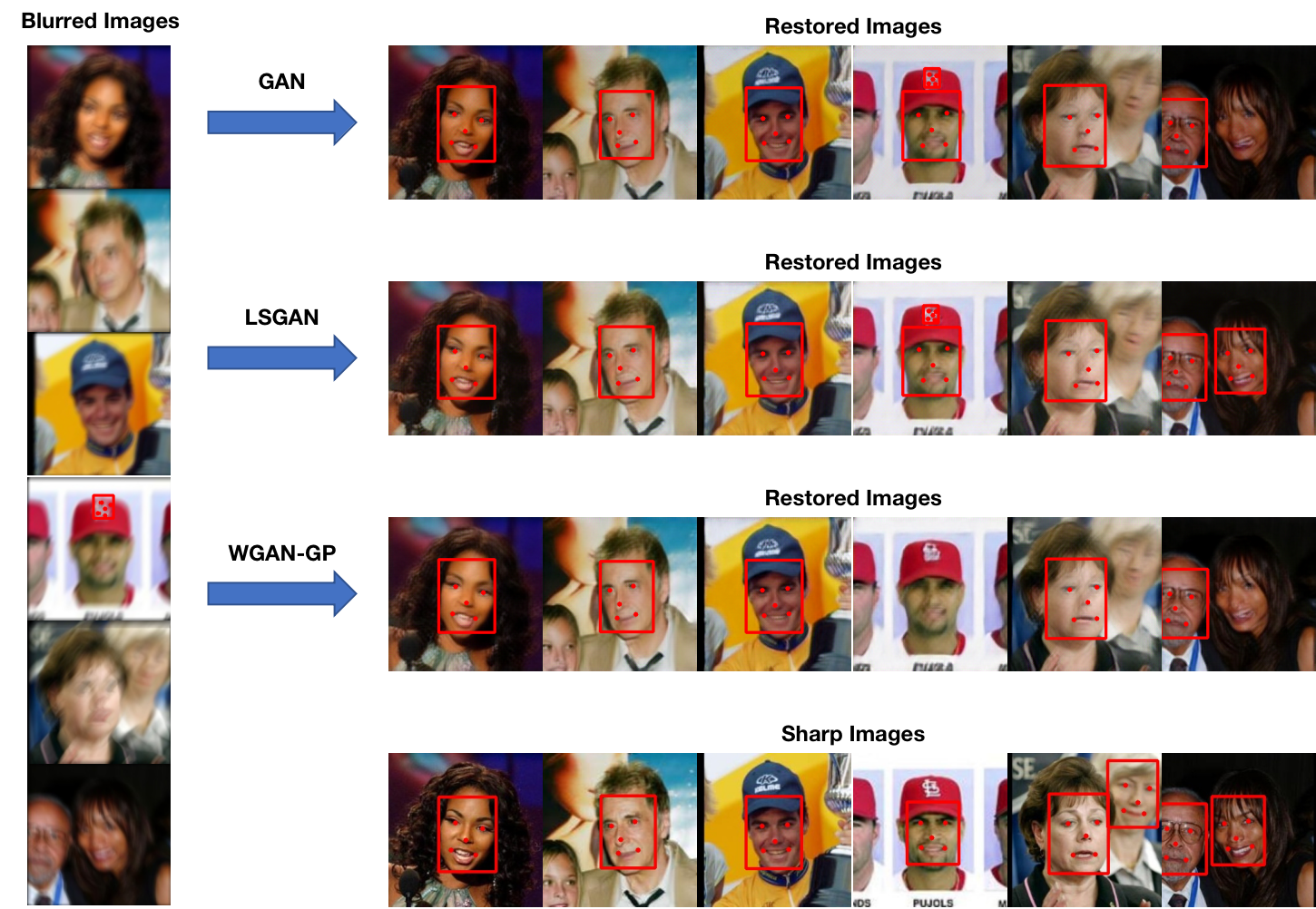}}
\caption{Deblurring processor under different GAN types help improve face detection results.}
\label{F:restored}
\end{figure}

We implemented three deblurring models under different types of GANs on PyTorch. The training used 1000 image pairs generated by the motion blur generation algorithm from the face images in LFW dataset, and we uniformly crop and resize the images to 256x256 during training. After the training is completed, we use other blurred face images that are also generated for the motion blur algorithm to perform deblurring tests, and put all sharp images, motion-blurred images, restored images into MTCNN \cite{zhang2016joint} for face detection comparison. From some of the results in Figure~\ref{F:restored}, we can see that the deblurring image processor not only restores the blur caused by motion visually, but also effectively helps MTCNN detect faces and the location of key-points.

As for the detailed face detection results, we conducted experiments on another 1,000 motion-blurred face image test sets under MTCNN, and compared the processed images with the unprocessed images as shown in the Table~\ref{tab:1}. The detection failure rate of pictures after the deblurring processor has dropped a lot under all the 3 types, which means that the recall rate has increased significantly. What’s more, the detection confidence of the faces is also increased.

\begin{table}[H]
    \caption{Motion-blurred faces detection results}
    \label{tab:1}
    \centerline{
    \begin{tabular}{ccc}
        \toprule
        Processor GAN Type       & Detection Failure Rate    & Detection Confidence Mean \\
        \midrule
        \textbf{GAN}             & \textbf{7.36\%}           & \textbf{0.9918}           \\
        \textbf{LSGAN}           & \textbf{6.61\%}           & \textbf{0.9916}           \\
        \textbf{WGAN-GP}         & \textbf{6.23\%}           & \textbf{0.9908}           \\
        Untreated blurred images & 21.9\%                    & 0.9729                    \\
        Sharp images             & 3.23\%                    & 0.9945                    \\
    \bottomrule
\end{tabular}
}

\vspace{.1in}
\end{table}

\section{Conclusion}\label{6}
This paper introduced an image processor for motion-blurred faces. It uses GAN structure, which can adapt well to blurred images without prior knowledge. From the face detection results of the processed face images, this kind of processor can also help face images perform more downstream tasks. In future work, we hope to improve the network structure and perform more targeted processing on face images. For example, when optimizing the network generator, we will consider adding the feature extraction and alignment of key points of the faces to the optimization goal to further improve the naturalness and restoration of the images generated by the GAN.

\bibliography{citation}

\begin{thebibliography}{10}

\bibitem{rowley1998neural}
Henry~A Rowley, Shumeet Baluja, and Takeo Kanade.
\newblock Neural network-based face detection.
\newblock {\em IEEE Transactions on pattern analysis and machine intelligence},
  20(1):23--38, 1998.

\bibitem{zhao2003face}
Wenyi Zhao, Rama Chellappa, P~Jonathon Phillips, and Azriel Rosenfeld.
\newblock Face recognition: A literature survey.
\newblock {\em ACM computing surveys (CSUR)}, 35(4):399--458, 2003.

\bibitem{yeh2017semantic}
Raymond~A Yeh, Chen Chen, Teck Yian~Lim, Alexander~G Schwing, Mark
  Hasegawa-Johnson, and Minh~N Do.
\newblock Semantic image inpainting with deep generative models.
\newblock In {\em Proceedings of the IEEE conference on computer vision and
  pattern recognition}, pages 5485--5493, 2017.

\bibitem{chen2018fsrnet}
Yu~Chen, Ying Tai, Xiaoming Liu, Chunhua Shen, and Jian Yang.
\newblock Fsrnet: End-to-end learning face super-resolution with facial priors.
\newblock In {\em Proceedings of the IEEE Conference on Computer Vision and
  Pattern Recognition}, pages 2492--2501, 2018.

\bibitem{zou2011very}
Wilman~WW Zou and Pong~C Yuen.
\newblock Very low resolution face recognition problem.
\newblock {\em IEEE Transactions on image processing}, 21(1):327--340, 2011.

\bibitem{yang2018proximal}
Dong Yang and Jian Sun.
\newblock Proximal dehaze-net: A prior learning-based deep network for single
  image dehazing.
\newblock In {\em Proceedings of the European Conference on Computer Vision
  (ECCV)}, pages 702--717, 2018.

\bibitem{goodfellow2014generative}
I~Goodfellow, J~Pouget-Abadie, M~Mirza, B~Xu, D~Warde-Farley, S~Ozair,
  A~Courville, and Y~Bengio.
\newblock Generative adversarial nets [c]//advances in neural information
  processing systems.
\newblock {\em New York: ACM}, 26722680, 2014.

\bibitem{antipov2017face}
Grigory Antipov, Moez Baccouche, and Jean-Luc Dugelay.
\newblock Face aging with conditional generative adversarial networks.
\newblock In {\em 2017 IEEE international conference on image processing
  (ICIP)}, pages 2089--2093. IEEE, 2017.

\bibitem{isola2017image}
Phillip Isola, Jun-Yan Zhu, Tinghui Zhou, and Alexei~A Efros.
\newblock Image-to-image translation with conditional adversarial networks.
\newblock In {\em Proceedings of the IEEE conference on computer vision and
  pattern recognition}, pages 1125--1134, 2017.

\bibitem{li2018beautygan}
Tingting Li, Ruihe Qian, Chao Dong, Si~Liu, Qiong Yan, Wenwu Zhu, and Liang
  Lin.
\newblock Beautygan: Instance-level facial makeup transfer with deep generative
  adversarial network.
\newblock In {\em Proceedings of the 26th ACM international conference on
  Multimedia}, pages 645--653, 2018.

\bibitem{fergus2006removing}
Rob Fergus, Barun Singh, Aaron Hertzmann, Sam~T Roweis, and William~T Freeman.
\newblock Removing camera shake from a single photograph.
\newblock In {\em ACM SIGGRAPH 2006 Papers}, pages 787--794. 2006.

\bibitem{xu2013unnatural}
Li~Xu, Shicheng Zheng, and Jiaya Jia.
\newblock Unnatural l0 sparse representation for natural image deblurring.
\newblock In {\em Proceedings of the IEEE conference on computer vision and
  pattern recognition}, pages 1107--1114, 2013.

\bibitem{xu2010two}
Li~Xu and Jiaya Jia.
\newblock Two-phase kernel estimation for robust motion deblurring.
\newblock In {\em European conference on computer vision}, pages 157--170.
  Springer, 2010.

\bibitem{perrone2014total}
Daniele Perrone and Paolo Favaro.
\newblock Total variation blind deconvolution: The devil is in the details.
\newblock In {\em Proceedings of the IEEE Conference on Computer Vision and
  Pattern Recognition}, pages 2909--2916, 2014.

\bibitem{babacan2012bayesian}
S~Derin Babacan, Rafael Molina, Minh~N Do, and Aggelos~K Katsaggelos.
\newblock Bayesian blind deconvolution with general sparse image priors.
\newblock In {\em European conference on computer vision}, pages 341--355.
  Springer, 2012.

\bibitem{gong2017motion}
Dong Gong, Jie Yang, Lingqiao Liu, Yanning Zhang, Ian Reid, Chunhua Shen, Anton
  Van Den~Hengel, and Qinfeng Shi.
\newblock From motion blur to motion flow: A deep learning solution for
  removing heterogeneous motion blur.
\newblock In {\em Proceedings of the IEEE conference on computer vision and
  pattern recognition}, pages 2319--2328, 2017.

\bibitem{sun2015learning}
Jian Sun, Wenfei Cao, Zongben Xu, and Jean Ponce.
\newblock Learning a convolutional neural network for non-uniform motion blur
  removal.
\newblock In {\em Proceedings of the IEEE Conference on Computer Vision and
  Pattern Recognition}, pages 769--777, 2015.

\bibitem{noroozi2017motion}
Mehdi Noroozi, Paramanand Chandramouli, and Paolo Favaro.
\newblock Motion deblurring in the wild.
\newblock In {\em German conference on pattern recognition}, pages 65--77.
  Springer, 2017.

\bibitem{nah2017deep}
Seungjun Nah, Tae Hyun~Kim, and Kyoung Mu~Lee.
\newblock Deep multi-scale convolutional neural network for dynamic scene
  deblurring.
\newblock In {\em Proceedings of the IEEE conference on computer vision and
  pattern recognition}, pages 3883--3891, 2017.

\bibitem{ramakrishnan2017deep}
Sainandan Ramakrishnan, Shubham Pachori, Aalok Gangopadhyay, and Shanmuganathan
  Raman.
\newblock Deep generative filter for motion deblurring.
\newblock In {\em Proceedings of the IEEE International Conference on Computer
  Vision Workshops}, pages 2993--3000, 2017.

\bibitem{huang2017densely}
Gao Huang, Zhuang Liu, Laurens Van Der~Maaten, and Kilian~Q Weinberger.
\newblock Densely connected convolutional networks.
\newblock In {\em Proceedings of the IEEE conference on computer vision and
  pattern recognition}, pages 4700--4708, 2017.

\bibitem{kupyn2018deblurgan}
Orest Kupyn, Volodymyr Budzan, Mykola Mykhailych, Dmytro Mishkin, and
  Ji{\v{r}}{\'\i} Matas.
\newblock Deblurgan: Blind motion deblurring using conditional adversarial
  networks.
\newblock In {\em Proceedings of the IEEE conference on computer vision and
  pattern recognition}, pages 8183--8192, 2018.

\bibitem{mao2017least}
Xudong Mao, Qing Li, Haoran Xie, Raymond~YK Lau, Zhen Wang, and Stephen
  Paul~Smolley.
\newblock Least squares generative adversarial networks.
\newblock In {\em Proceedings of the IEEE international conference on computer
  vision}, pages 2794--2802, 2017.

\bibitem{arjovsky2017wasserstein}
Martin Arjovsky, Soumith Chintala, and L{\'e}on Bottou.
\newblock Wasserstein generative adversarial networks.
\newblock In {\em International conference on machine learning}, pages
  214--223. PMLR, 2017.

\bibitem{gulrajani2017improved}
Ishaan Gulrajani, Faruk Ahmed, Martin Arjovsky, Vincent Dumoulin, and Aaron
  Courville.
\newblock Improved training of wasserstein gans.
\newblock {\em arXiv preprint arXiv:1704.00028}, 2017.

\bibitem{villani2008optimal}
C{\'e}dric Villani.
\newblock {\em Optimal transport: old and new}, volume 338.
\newblock Springer Science \& Business Media, 2008.

\bibitem{boracchi2012modeling}
Giacomo Boracchi and Alessandro Foi.
\newblock Modeling the performance of image restoration from motion blur.
\newblock {\em IEEE Transactions on Image Processing}, 21(8):3502--3517, 2012.

\bibitem{he2016deep}
Kaiming He, Xiangyu Zhang, Shaoqing Ren, and Jian Sun.
\newblock Deep residual learning for image recognition.
\newblock In {\em Proceedings of the IEEE conference on computer vision and
  pattern recognition}, pages 770--778, 2016.

\bibitem{simonyan2014very}
Karen Simonyan and Andrew Zisserman.
\newblock Very deep convolutional networks for large-scale image recognition.
\newblock {\em arXiv preprint arXiv:1409.1556}, 2014.

\bibitem{huang2008labeled}
Gary~B Huang, Marwan Mattar, Tamara Berg, and Eric Learned-Miller.
\newblock Labeled faces in the wild: A database forstudying face recognition in
  unconstrained environments.
\newblock In {\em Workshop on faces in'Real-Life'Images: detection, alignment,
  and recognition}, 2008.

\bibitem{zhang2016joint}
Kaipeng Zhang, Zhanpeng Zhang, Zhifeng Li, and Yu~Qiao.
\newblock Joint face detection and alignment using multitask cascaded
  convolutional networks.
\newblock {\em IEEE Signal Processing Letters}, 23(10):1499--1503, 2016.

\end{thebibliography}

\end{document}